%% file: main.tex
\documentclass[10pt,twocolumn,letterpaper]{article}

\usepackage[pagenumbers]{cvpr} 

\input{preamble}

\definecolor{cvprblue}{rgb}{0.21,0.49,0.74}
\definecolor{ZouColor}{rgb}{1.0,0.0,1.0} 
\definecolor{ZouColor2}{rgb}{1.0,0.0,0.0} %

\usepackage[pagebackref,breaklinks,colorlinks,citecolor=cvprblue]{hyperref}

\def\paperID{6295} 
\def\confName{CVPR}
\def\confYear{2024}

\title{3D-SceneDreamer: Text-Driven 3D-Consistent Scene Generation}

\author{Songchun Zhang$^1$, Yibo Zhang$^2$, Quan Zheng$^4$, Rui Ma$^2$, Wei Hua$^3$ \\Hujun Bao$^1$, Weiwei Xu$^1$, Changqing Zou$^{1,3}$ \\
[2mm]
$^1$~Zhejiang University \quad $^2$ Jilin University \quad $^3$~Zhejiang Lab\\ 
$^4$~Institute of Software, Chinese Academy of Sciences
}

\begin{document}
\maketitle
\vspace{-2mm}
\input{figure/first_teaser}

\input{sec/0_abstract}    
\input{sec/1_intro}

\input{sec/2_related}
\input{sec/3_preliminaries}
\input{sec/4_methods}

\input{sec/5_experiments}

\input{sec/6_conclution}

{
    \small
    \bibliographystyle{ieeenat_fullname}
    \bibliography{main}
}


\end{document}

%% file: preamble.tex
\usepackage{booktabs}
\usepackage{multirow}
\usepackage[table,xcdraw]{xcolor}
\usepackage{cuted}
\usepackage{soul} 
\usepackage{amssymb}
\usepackage{pifont}

\newcommand{\todo}[1]{{\color{red}#1}}


%% file: figure/first_teaser.tex
\begin{strip}
    \centering
    \vspace{-5em}
    \includegraphics[width=1.0\textwidth]{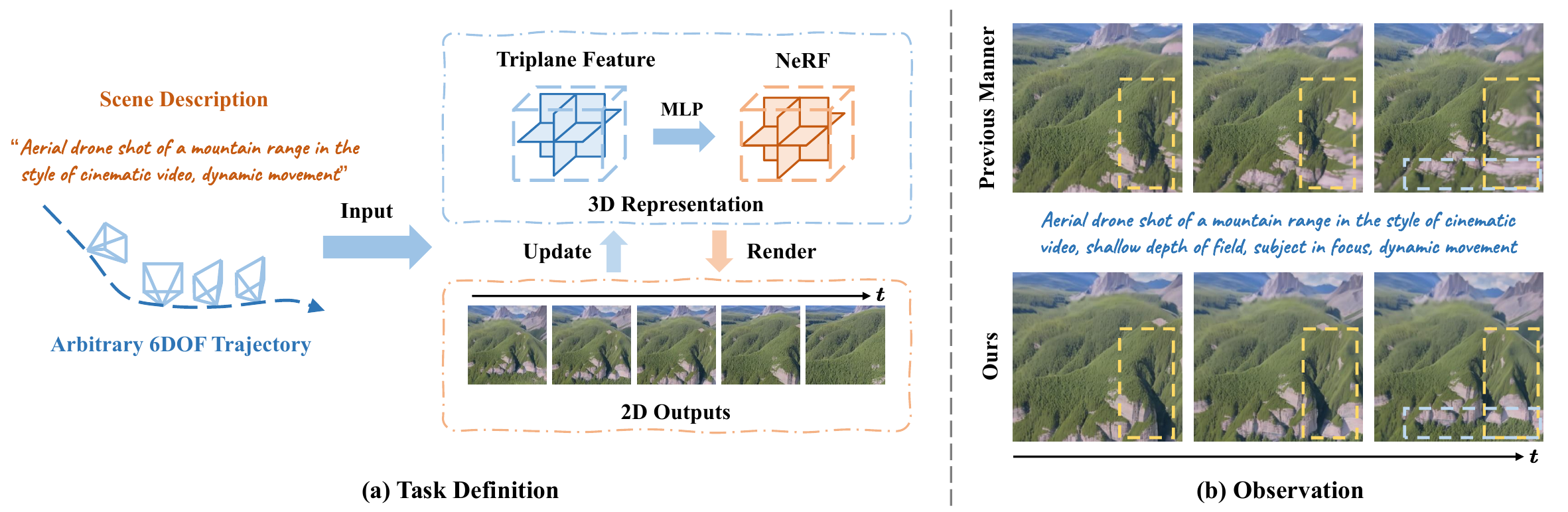}
    \vspace{-2em}
     \captionof{figure}{\textbf{Text-Driven 3D Scene Generation from text prompts.} (a) Given a scene description prompt and an arbitrary 6-degree-of-freedom (6-DOF) camera trajectory, our approach progressively generates the full 3D scene by continuously synthesizing 2D novel views. (b) The limitation of mesh representations~\cite{text2room, scenescape} and the lack of reasonable rectification mechanisms lead to cumulative errors in outdoor scenes, which are respectively marked with yellow and blue dash line boxes.
     In contrast, our approach can alleviate the problem by introducing a progressive generation pipeline.}
     \vspace{-1em}
    \label{fig:first_teaser}
\end{strip}

%% file: sec/0_abstract.tex
\begin{abstract}
\vspace{-4mm}
Text-driven 3D scene generation techniques have made rapid progress in recent years.
Their success is mainly attributed to using existing
generative models to iteratively perform image warping and inpainting to generate 3D scenes.
However, these methods heavily rely on the outputs of existing models, leading to error accumulation in geometry and appearance that prevent the models from being used in various scenarios (e.g., outdoor and unreal scenarios).
To address this limitation, we generatively refine the newly generated local views by querying and aggregating global 3D information, and then progressively generate the 3D scene.
Specifically, we employ a tri-plane features-based NeRF as a unified representation of the 3D scene to constrain global 3D consistency, and propose a generative refinement network to synthesize new contents with higher quality by exploiting the natural image prior from 2D diffusion model as well as the global 3D information of the current scene. 
Our extensive experiments demonstrate that, in comparison to previous methods, our approach supports wide variety of scene generation and arbitrary camera trajectories with improved visual quality and 3D consistency.
\end{abstract}

%% file: sec/1_intro.tex
\vspace{-4mm}
\section{Introduction}
\label{sec:intro}

\indent

In recent years, with the growing need for 3D creation tools for metaverse applications, attention to 3D scene generation techniques has increased rapidly.
Existing tools~\cite{infigen, blender} usually require professional modeling skills and extensive manual labor, which is time-consuming and inefficient.
To facilitate the 3D scene creation and reduce the need for professional skills, 3D scene generation tools should be intuitive and versatile while ensuring sufficient controllability.

\input{figure/teaser}
\input{table/method_compare}
This paper focuses on the specific setting of generating consistent 3D scenes from the input texts that describe the 3D scenes.
This problem is highly challenging from several perspectives, including the limitation of available text-3D data pairs and the need for ensuring both semantic and geometric consistency of the generated scenes.
To overcome the limited 3D data issue, recent text-to-3D methods~\cite{dreamfusion, Prolificdreamer} have leveraged the powerful pre-trained text-to-image diffusion model~\cite{stablediffusion} as a strong prior to optimize 3D representation.
However, their generated scenes often have relatively simpler geometry and lack 3D consistency, because 2D prior diffusion models lack the perception of 3D information. 

Some recent methods~\cite{scenescape, text2room} introduce the monocular depth estimation model~\cite{dpt, midas} as a strong geometric prior and follow the \textit{warping-inpainting} pipeline~\cite{liu2021infinite, Infinitenature-zero} for progressive 3D scene generation, which partially solves the inconsistency problem.
Although these methods can generate realistic scenes with multi-view 3D consistency, they mainly focus on indoor scenes and fail to handle large-scale outdoor scene generation as illustrated in~\cref{fig:first_teaser}~(b).
This can be attributed to two main aspects: (1) Due to the adoption of an explicit 3D mesh as the unified 3D representation, the noise of the depth estimation in the outdoor scene can cause a large stretch of the scene geometry; (2) The lack of an efficient rectification mechanism in the pipeline leads to an accumulation of geometric and appearance errors.

In this paper, we present a new framework, named \textbf{3D-SceneDreamer} that provides a unified solution for text-driven 3D consistent indoor and outdoor scene generation.
Our approach employs a tri-planar feature-based radiance field as a unified 3D representation instead of 3D mesh, which is advantageous for general scene generation (especially in outdoor scenes) and supports navigating with arbitrary 6-DOF camera trajectories.
Afterwards, we model the scene generation process as a progressive optimization of the NeRF representation, while a text-guided and scene-adapted generative novel view synthesis is employed to refine the NeRF optimization.
~\cref{fig:teaser} shows a comparison of our design with existing text-to-scene pipelines.

Specifically, we first perform scene initialization, which consists of two stages, i.e., generating a supporting database and optimizing the initial scene representation.
We first use the input text prompt and the pre-trained diffusion model ~\cite{stablediffusion} to generate the initial image as an appearance prior. 
Then, we use an off-the-shelf depth estimation model~\cite{bhat2023zoedepth} to provide the geometric prior for the corresponding scene.
Inspired by~\cite{text2nerf}, to prevent NeRF from over-fitting for the single view image, we construct a database via differentiable spatial transformation~\cite{jaderberg2015spatial} and use it for optimizing the initial NeRF representation of the generated scene.
To generate the extrapolated content, we use volume rendering and trilinear interpolation in the novel viewpoints to obtain the initial rendered images and their corresponding feature maps.
These outputs are later fed into our 3D-aware generative refinement model, whose output images are subsequently added as new content to the supporting database.
Next, in conjunction with the new data, we progressively generate the whole 3D scene by updating our 3D representation through our incremental training strategy.

Extensive experiments demonstrate that our approach significantly outperforms the state-of-the-art text-driven 3D scene generation method in both visual quality and 3D consistency.
To summarize, our technical contributions are as follows:
\begin{itemize}
    \item We provide a unified solution for text-driven consistent 3D scene generation that supports both indoor and outdoor scenes as well as allows navigation with arbitrary 6-DOF camera trajectories.
    \item We propose to use a tri-planar feature-based neural radiance field as a global 3D representation of the scene to generate continuous scene views, which preserves the 3D consistency of the scene, empowered by a progressive optimization strategy.
    \item We propose a new generative refinement model, which explicitly injects 3D information to refine the coarse view generated by novel view synthesis and then incorporates the new views to refine the NeRF optimization.
\end{itemize}


%% file: figure/teaser.tex
\begin{figure}[t!]
    \centering
    \includegraphics[width=0.48\textwidth]{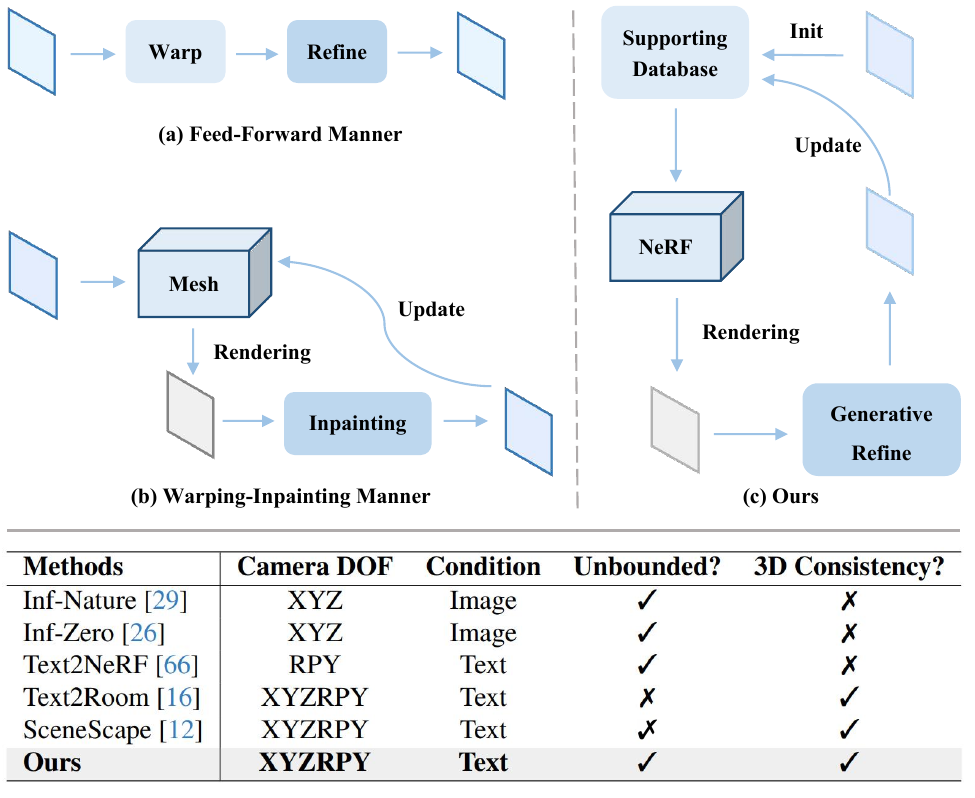}
    \vspace{-6mm}
    \caption{\textbf{Comparison with existing designs.} (a) The feed-forward approaches use depth-based warping and refinement operations to generate novel views of the scene without a unified representation. (b) The warping-inpainting approaches use mesh as a unified representation and generate the scene through iterative inpainting. (c) We replace the mesh with NeRF as the unified representation and alleviate the cumulative error issue by incorporating a generative refinement model. This allows our framework to support the generation of a wider range of scene types. The table at the bottom illustrates the unique feature of the proposed approach. We use a tick with a cross on it for SceneScape because it only supports backward camera movement, not able to provide a full unbounded generation.}
    \label{fig:teaser}
    \vspace{-6mm}
\end{figure}

%% file: table/method_compare.tex


%% file: sec/2_related.tex
\section{Related Work}
\label{sec:related_work}


\noindent{\textbf{Text-Driven 3D Content Generation.}}
%
Recently, motivated by the success of text-to-image models, employing pre-trained 2D diffusion models to perform text-to-3D generation has gained significant research attention. 
Some pioneering works~\cite{dreamfusion,SJC} introduce the Score Distillation Sampling (SDS) and utilize 2D diffusion prior to optimize 3D representation. 
Subsequent works~\cite{Latent-NeRF,magic3d,Fantasia3D,Prolificdreamer} further enhance texture realism and geometric quality. However, they primarily focus on improving object-level 3D content generation rather than large-scale 3D scenes. 
Recent works~\cite{scenescape,text2nerf,text2room} have proposed some feasible solutions for 3D scene generation.
By utilizing the pre-trained monocular depth model and the inpainting model, they generate the 3D scene progressively based on the input text and camera trajectory.
However, due to the underlying 3D representation or optimization scheme, these methods are limited in several aspects.
For example, as \cite{scenescape, text2room} utilize explicit mesh as 3D representation, it is difficult for them to generate outdoor scenes. 
Besides, their mesh outputs also suffer from fragmented geometry and artifacts due to imprecise depth estimation results.
Although Text2NeRF achieves to generate high-quality indoor and outdoor scenes by replacing the meshes with neural radiance fields \cite{NeRF}, it can only generate camera-centric scenes.
In contrast, our approach not only supports more general 3D scene generation but can also handle arbitrary 6DOF camera trajectories.

\noindent{\textbf{Text-Driven Video Generation.}}
Text-Driven Video Generation aims to create realistic video content based on textual conditions. 
In the early stages, this task was approached using GAN \cite{balaji2019conditional,pan2017create,li2018video} and VAE  \cite{mittal2017sync,marwah2017attentive} generative models, but the results were limited to low-resolution short video clips.
Following the significant advancements in text-to-image models, recent text-to-video works extend text-to-image models such as transformer \cite{wu2021godiva, cogvideo, wu2022nuwa} and diffusion model \cite{Video-diffusion-models,Make-a-video,ho2022imagen,VideoFusion,Align-your-Latents,magicvideo} for video generation.
These approaches enable the generalization of high-quality and open-vocabulary videos, but require a substantial amount of text-image or text-video pairs of data for training. 
Text2Video-Zero \cite{Text2video-zero} proposes the first zero-shot text-to-video generation pipeline that does not rely on training or optimization, but their generated videos lack smoothness and 3D consistency. 
Our method is capable of generating smooth and long videos wihch are consistent to the scenes described by the input text, without the need for large-scale training data. 
Furthermore, the utilization of NeRF as the 3D representation enhances the 3D consistency of our videos.

\noindent{\textbf{View Synthesis with Generative Models.}}
Several early stage studies \cite{liu2021infinite, Infinitenature-zero, koh2021pathdreamer, wiles2020synsin, koh2023simple, chan2021pi} employ GAN to synthesize new viewpoints. 
However, the training process of GAN is prone to the issue of mode collapse, which limits the diversity of generation results. 
Diffusion model has been shown its capability to generate diverse and high-quality images and videos. 
In recent view synthesis works \cite{tseng2023consistent, chan2023generative, Cai2022DiffDreamerCS, Make-It-4D}, diffusion models have been employed to achieve improved scene generation results over prior works.
For example, in Deceptive-NeRF \cite{liu2023deceptive}, pseudo-observations are synthesized by diffusion models and these observations are further utilized for enhance the NeRF optimization. 
Closely similar to \cite{liu2023deceptive}, our method propose a geometry-aware diffusion refinement model to reduce the artifacts of the input coarse view generated by the initial novel view synthesis.
With the 3D information from NeRF features injected to the refinement process, we can achieve globally consistent 3D scene generation.





%% file: sec/3_preliminaries.tex
\section{Neural Radiance Fields Revisited}
\label{sec:preliminaries}

Neural Radiance Fields (NeRF)~\cite{wang2022nerf} is a novel view synthesis technique that has shown impressive results.
It represents the specific 3D scene via an implicit function, denoted as $f_\theta:(\boldsymbol{x}, \boldsymbol{d}) \mapsto (\mathbf{c}, \sigma)$, given a spatial location $\mathbf{x}$ and a ray direction $\mathbf{d}$, where $\theta$ represents the learnable parameters, and $\mathbf{c}$ and $\sigma$ are the color and density.
To render a novel image, NeRF marches a camera ray $\mathbf{r}(t) = \mathbf{o} + t\mathbf{d}$ starting from the origin $\mathbf{o}$ through each pixel and calculates its color $\hat{\boldsymbol{C}}$ and rendered depth $\hat{\boldsymbol{D}}$ via the volume rendering quadrature, \ie, $\hat{\boldsymbol{C}}(\mathbf{r})=\sum_{i=1}^N T_i \alpha_i \mathbf{c}_i$ and $\hat{\boldsymbol{D}}(\mathbf{r})=\sum_{i=1}^N T_i \alpha_i t_i$,
%
\noindent
where $T_i=\exp \left(-\sum_{j=1}^{i-1} \sigma_j \delta_j\right)$, $\alpha_i = \left(1-\exp \left(-\sigma_i \delta_i\right)\right)$, and $\delta_{k}=t_{k+1}-t_k$ indicates the distance between two point samples. 
Typically, stratified sampling is used to select the point samples $\{t_i\}_{i=1}^N$ between $t_n$ and $t_f$, which denote the near and far planes of the camera.
When multi-view images are available, $\theta$ can be easily optimized with the MSE loss:
\vspace{-3mm}
\begin{equation}
\mathcal{L}_\theta=\sum_{\boldsymbol{r} \in \mathcal{R}}\left\|\boldsymbol{\hat{C}}(\boldsymbol{r})-\boldsymbol{C}(\boldsymbol{r})\right\|_2^2
\end{equation} 
\vspace{-3mm}

\noindent 
where $\mathcal{R}$ is the collection of rays, and $\boldsymbol{C}$ indicates the ground truth color.

%% file: sec/4_methods.tex
\section{Methods}
\label{sec:methods}

\subsection{Overview}
\input{figure/pipeline}
Given a description of the target scene a the input text prompt $\mathbf{p}$, and a pre-defined camera trajectory denoted by $\{\mathbf{T}_i\}_{i=1}^N$, our goal is to generate a 3D scene along the camera trajectory with the multiview 3D consistency.


The overview of the proposed model is illustrated in~\cref{fig:pipeline}.
We first introduce the acquisition of appearance and structural priors in~\cref{sec:initialization}, which serve as the scene initialization.
The formulation of Unified Scene Representation and its optimization with the former priors are presented in~\cref{sec:Representation}.
To synthesize new content while maintaining the multiview consistency, we propose a geometry-aware refinement model in~\cref{sec:refinement}.
Finally, the full online scene generation process is presented in~\cref{sec:Online}.

\subsection{Scene Context Initialization}
\label{sec:initialization}
Given the input textual prompt $\mathbf{p}$, we first utilize a pre-trained stable diffusion model to generate an initial 2D image $\mathbf{I}_0$, which serves as an appearance prior for the scene. 
Then, we feed this image into the off-the-shelf depth estimation model~\cite{bhat2023zoedepth}, and take the output as a geometric prior for the target scene, denoted as $\mathbf{D}_0$.
Inspired by~\cite{text2nerf}, we construct a supporting database $\mathcal{S}=\{(\mathbf{D}_i, \mathbf{I}_i, \mathbf{T}_i)\}_{i=1}^N$ via differentiable spatial transformation~\cite{jaderberg2015spatial} and image inpainting~\cite{text2room} techniques, where $N$ denotes the number of initial viewpoints.
This database provides additional views and depth information, which could prevent the model from overfitting to the initial view.
With the initial supporting database, we can initialize the global 3D representation.
The data generated by our method will be continuously appended to this supporting database for continuous optimization of the global 3D representation.
More details are provided in our supplemental materials.

\subsection{Unified Scene Representation}
\label{sec:Representation}
%
Though previous methods~\cite{Infinitenature-zero, liu2021infinite} have achieved novel view generations via differentiable rendering-based frame-to-frame warping, there are still drawbacks: (1) the global 3D consistency is not ensured, (2) cumulative errors occur in long-term generation, (3) complex scenes may lead to failure.
To tackling above issues, we propose a tri-planar feature-based NeRF as the unified representation. 
Compared with previous methods~\cite{Infinitenature-zero, liu2021infinite,text2room, scenescape}, our approach constrains the global 3D consistency while handling the scene generation with complex appearances and geometries.

\noindent{\textbf{Tri-planar Feature Representation.}
For constructing the feature tri-planes $\mathbf{M}=\{\mathbf{M}_{xy}, \mathbf{M}_{yz}, \mathbf{M}_{xz}\} \in \mathbb{R}^{3 \times S \times S \times D}$ from the input images, where $S$ is the spatial resolution and $D$ is the feature dimension, we first extract 2D image features from supporting views using the pre-trained ViT from DINOv2~\cite{oquab2023dinov2} because of its strong capability in modeling cross-view correlations.
We denote the extracted feature corresponding to image $\mathbf{I}_i$ as $\mathbf{F}_i$, and the feature set obtained from all input views is denoted as $\{\mathbf{F}_i\}_{i=1}^N$.
To lift the local 2D feature maps into the unified 3D space, similar to the previous work~\cite{zhang2022nerfusion}, we back-project the extracted local image features $\mathbf{F}$ into a 3D feature volume $\mathbf{V}$ along each camera ray.
To avoid the cubic computational complexity of volumes, we construct a tri-planar representation by projecting the 3D feature volume $\mathbf{V}$ into its respective plane via three separate encoders.
This representation reduces the complexity from feature dimensionality reduction, but with equivalent information compared to purely 2D feature representations (e.g., BEV representations~\cite{chen2023scenedreamer,li2022bevformer}).

\todo{}

\noindent{\textbf{Implicit Radiance Field Decoder.}}
Based on the constructed tri-planar representation $\mathbf{M}$, we can reconstruct the images with target poses via our implicit radiance field decoder module $\Psi=\{f_g,f_c\}$, where $f_g$ and $f_c$ indicate the geometric feature decoder and appearance decoder.
Given a 3D point $p = [i,j,k]$ and a view direction $\boldsymbol{d}$, we orthogonally project $p$ to each feature plane in $\mathbf{M}$ with bilinear sampling to obtain the conditional feature $\mathbf{M}_p=[\textbf{M}_{xy}(i,j),\textbf{M}_{yz}(j,k),\textbf{M}_{xz}(i,k)]$.
We feed $\mathbf{M}_p$ into the geometric feature decoder to obtain the predicted density $\sigma$ and the geometric feature vector $\boldsymbol{g}$, after which we further decode its color $\boldsymbol{c}$: 

\begin{equation}
\begin{gathered}
(\sigma, \boldsymbol{g})=f_g\left(\gamma(\boldsymbol{x}), \mathbf{M}_p\right) \\
\boldsymbol{c}=f_c\left(\gamma(\boldsymbol{x}), \gamma(\boldsymbol{d}), \boldsymbol{g}, \mathbf{M}_p\right)
\end{gathered}
\end{equation}
where $\gamma(\cdot)$ indicates the positional encoding function.
Then we can calculate the pixel color via an approximation of the volume rendering integral mentioned in~\cref{sec:preliminaries}.

\par
\noindent{\textbf{Training Objective.}}
To optimize our 3D representation, we leverage the ground truth colors from the target image as the supervisory signal.
Additionally, in the setting with sparse input views, we employ the estimated dense depth map to enhance the model's learning of low-frequency geometric information and prevent overfitting to appearance details. 
Our optimization objective is as follows:
\vspace{-1.5mm}
\begin{equation}\label{eq:obj}
\mathcal{L}=\sum_{\boldsymbol{r}\in\mathcal{R}}\left(\mathcal{L}_{photo}\left(\boldsymbol{r}\right)+\lambda\mathcal{L}_{depth}\left(\boldsymbol{r}\right)\right)
\end{equation} 

\vspace{-2.5mm}
\noindent where $\mathcal{L}_{photo}\left(\boldsymbol{r}\right)=\left\Vert\hat{\boldsymbol{C}}\left(\boldsymbol{r}\right)-\boldsymbol{C}\left(\boldsymbol{r}\right)\right\Vert^{2}$, $\mathcal{L}_{depth}\left(\boldsymbol{r}\right)=\left\|\mathbf{\hat{D}^*_r}\left(\boldsymbol{r}\right)-\mathbf{D^*}\left(\boldsymbol{r}\right)\right\|^2$,
$\mathcal{R}$ denotes the collection of rays generated from the images in the supporting database, $\lambda$ indicates the balance weight of the depth loss, and $\mathbf{D^*}(\boldsymbol{r})$ and $\mathbf{\hat{D}^*_r}(\boldsymbol{r})$ denote the rendered depth and the depth obtained from the pre-trained depth estimation model. 
Since monocular depths are not scale- and shift-invariant, both depths are normalized per frame.


\subsection{3D-Aware Generative Refinement}
\label{sec:refinement}
Given a sequence of poses and an initial viewpoint, previous methods~\cite{text2nerf, scenescape, text2room} usually generate novel views by the \textit{warping-inpainting} pipeline.
Though these methods have achieved promising results, they suffer from two issues:
(1) The lack of rectification mechanisms in these methods can lead to error accumulation.
(2) The lack of explicit 3D information during the inpainting process of these methods can lead to insufficient 3D consistency.
Therefore, we propose a 3D-Aware Generative Refinement model to alleviate the above issues.
On the one hand, we introduce an efficient refinement mechanism to reduce the cumulative error in the novel view generation.
On the other hand, we explicitly inject 3D information during the process of generating novel views to enhance 3D consistency.
We will describe the model design below.
\par
\noindent{\textbf{Model Design.}}
Given a novel viewpoint with camera pose $\mathbf{T}_i$, the tri-planar features $\mathbf{M}$, we can obtain the rendered image $\mathbf{I}_r$, rendered depth $\mathbf{D}_r$ and the corresponding 2D feature map $\mathbf{F}_r$ via the radiance field decoder module $\Psi$ and volume rendering.
For convenience, we model the whole process with a mapping operator $
\mathcal{F}_{ren}: \{\mathbf{T}_i, \mathbf{M}\} \mapsto \{\mathbf{I}_r, \mathbf{F}_r, \mathbf{D}_r\}$.
Note that the feature map is computed similarly to the color and depth, \ie, by numerical quadrature, and can be formulated as
\vspace{-1.5mm}
\begin{equation}
{\mathbf{F}}_r(\mathbf{r})=\sum_{i=1}^N T_i\left(1-\exp \left(-\sigma_i \delta_i\right)\right) \boldsymbol{g}_i
\end{equation}

\vspace{-1mm}
\noindent where $\boldsymbol{g}_i$ indicates the feature vector decoded by $f_g$, and $N$ denotes the total number of point samples on the ray $\boldsymbol{r}$.


Although the quality of the rendered coarse results may not be very high, they can still provide reasonable guidance for the extrapolated view generation according to the current scene.
Based on this assumption, we propose to take the rendered image and the feature map as conditional inputs to a pre-trained 2D stable diffusion model and generate a refined synthetic image $\hat{\mathbf{I}}_r$ via fine-tuning the model, which allows to leverage natural image priors derived from internet-scale data.
The process can be formulated as:
\begin{equation}
    \hat{\mathbf{I}}_r = \mathcal{F}_{gen}(\mathbf{I}_r, \tau(\mathbf{p}), \mathcal{G}(\mathbf{F}_r))
\end{equation}
where $\mathcal{F}_{gen}$ denotes our generative refinement model, $ \tau(\mathbf{p})$ indicates the input text embedding, and $\mathcal{G}$ denotes the feature adapter for learning the mapping from external control information to the internal knowledge in LDM.

\noindent{\textbf{Scene-Adapted Diffusion Model Fine-Tuning.}}
For the scene generation task, we propose to leverage the rich 2D priors in the pre-trained latent diffusion model instead of training a new model from scratch.
Thus, we jointly train the feature adapter, the radiance field decoder, and the feature aggregation layer, while keeping the parameters of stable diffusion fixed.
The objective of the fine-tuning process is shown below:

\vspace{-4mm}
\begin{equation}
\label{eq:ad}
     \mathcal{L}_{A D}=\mathbb{E}_{t, \epsilon \sim \mathcal{N}(0, I)}\left[\left\|\epsilon_\theta\left(\boldsymbol{z}_t, t, \tau(\mathbf{p}), \mathbf{F}_r, \mathbf{I}_r\right)-\epsilon\right\|_2^2\right]
\end{equation}

With the rendered feature map $\mathbf{F}_r$ containing information about the appearance and geometry, we can control the pre-trained text-to-image diffusion model to generate images that are consistent with the content of generated images from previous viewpoints. 
In addition, our model inherits the high-quality image generation ability of the stable diffusion model, which ensures the plausibility of the generated views. 
The pre-trained prior and our effective conditional adaptation enable our model to have generalization ability in novel scenes.
\par
\noindent{\textbf{Global-Local Consistency Regularization.}}
In the online generation process, though our model can rectify the coarse rendering results, we do not explicitly constrain the 3D consistency across views when synthesizing novel views.
Therefore, we design a regularization term $\mathcal{L}_{cons}$ for test-time optimization, which shares the same formula as~\cref{eq:ad} to guarantee the plausibility of the generated novel views.
Specifically, we expect that 3D consistency exists between novel views obtained from geometric projection using local geometric information (i.e., monocular depth estimation) and novel views generated using global geometric information (i.e., global tri-planar 3D representation).
Thus, we simultaneously generate novel views based on the previous warping-and-inpainting pipeline and use them as supervisory signals to further fine-tune the feature adapter.

\subsection{Online Scene Generation Process.}
\label{sec:Online}
In this section, we introduce our online 3D scene generation process, which consists of three parts: scene representation initialization, extrapolation content synthesis, and incremental training strategy.
\par
\noindent{\textbf{Scene Representation Initialization.}}
Given the input textual prompt, we first generate an initial 2D image using a pre-trained stable diffusion model, after which we construct a supporting database $\mathcal{S}$ via the method mentioned in~\cref{sec:initialization}.
Then, by exploiting the data from the database, as well as the photometric loss (\cref{eq:obj}), we can optimize the unified representation.
To prevent the model from overfitting to high-frequency details, we allow the model to learn low-frequency geometric information better by utilizing the depth priors.~\cite{wang2023sparsenerf}.
\par
\noindent{\textbf{Extrapolated Content Synthesis.}}
To generate the extrapolated content, we proceed by retrieving the next pose, denoted as $\mathbf{T}_i$, from the pose sequence $\{\mathbf{T}_i\}_{i=1}^N$. 
We then employ volumetric rendering to obtain a coarse view of the current viewpoint and the corresponding feature map.
These rendered outputs are used as conditional inputs to our generative refinement model $\mathcal{F}_{gen}$ for generating a refined view.
Due to the presence of a generative refinement mechanism, our extrapolation method mitigates the effects of cumulative errors. 
The refined view from the model $\mathcal{F}_{gen}$ is subsequently added to the supporting database $\mathcal{S}$ as new content.
\par
\noindent{\textbf{Incremental Training Strategy.}}
After obtaining the new content, we then need to update the unified representation. 
However, fine-tuning only on the newly generated data can lead to catastrophic forgetting, whereas fine-tuning on the entire dataset requires excessively long training time.
Inspired by~\cite{sucar2021imap}, we sample a sparse set of rays $\mathcal{Q}$ according to the information gain to optimize the representation, thus improving the efficiency of the incremental training.


%% file: figure/pipeline.tex
\begin{figure*}[t!]
    \centering
    \includegraphics[width=1.0\textwidth]{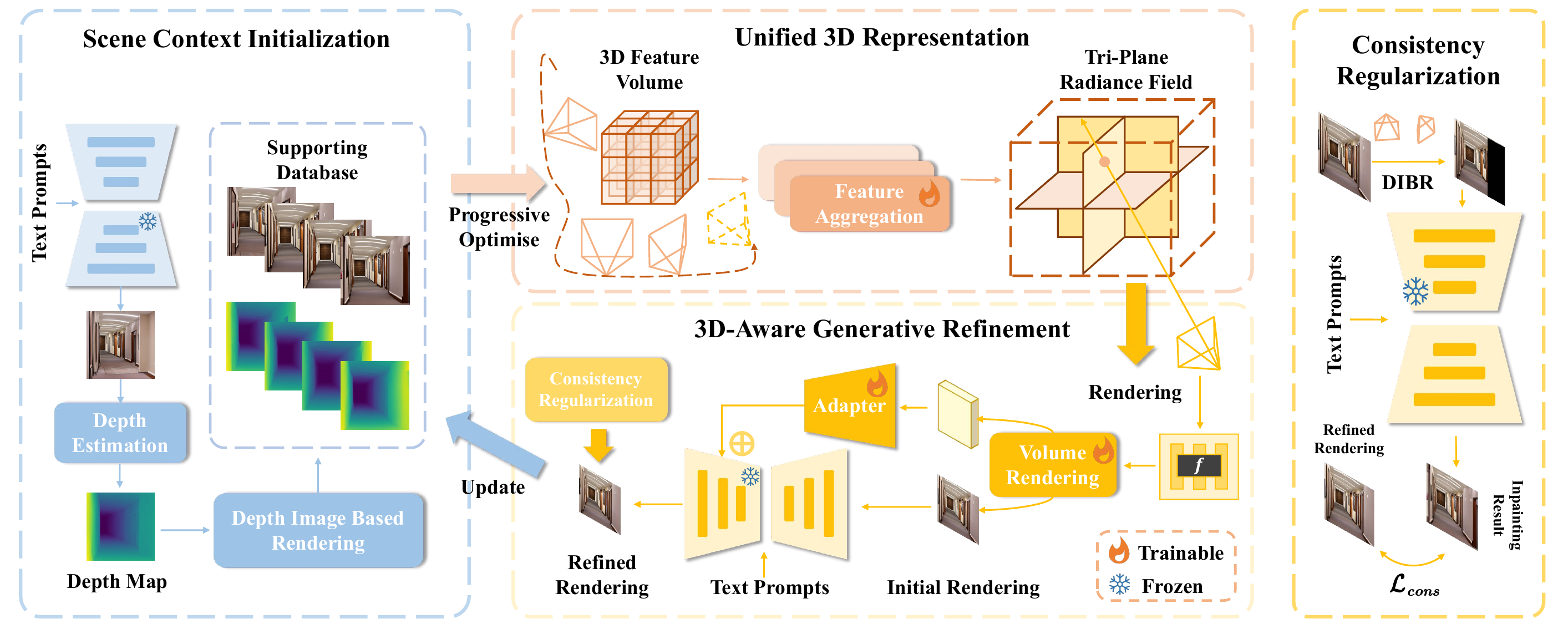}
    \vspace{-6mm}
    \caption{\textbf{Overview of our pipeline.} (a) \textbf{Scene Context Initialization} contains a supporting database to provide novel viewpoint data for progressive generation. (b) \textbf{Unified 3D Representation} provides a unified representation for the generated scene, which allows our approach to accomplish more general scene generation and to hold the 3D consistency at the same time. (c) \textbf{3D-Aware Generative Refinement} alleviates the cumulative error issue during long-term extrapolation by exploiting large-scale natural images prior to generatively refine the synthesized novel viewpoint image. The consistency regularization module is used for test-time optimization.}
    \label{fig:pipeline}
    \vspace{-6mm}
\end{figure*}

%% file: sec/5_experiments.tex
\section{Experiments}
\label{sec:experiments}

\subsection{Implementation details.}

\input{figure/figure_result}
\input{figure/text2360}
We implemented our system using PyTorch.
For the differentiable rendering part, we utilized~\cite{guizilini2023towards} for depth estimation.
To avoid the occurrence of black holes, we referred to the implementation in~\cite{jaderberg2015spatial} to generate surrounding views.
For the text-guided image generation, we use the publicly available stable diffusion code from Diffusers~\cite{von-platen-etal-2022-diffusers}.
For the multi-view consistency image generation, we refer to the implementation of T2I-Adapter~\cite{mou2023t2i} to inject the depth feature conditions.
In the progressive NeRF reconstruction part, we refer to the tri-planar implementation in~\cite{eg3d}.
We conducted all experiments using 4 NVIDIA RTX A100 GPUs for training and inference.
More details can be found in our supplementary material.
\par
\subsection{Evaluation metrics.}
\label{sec:evaluation_metrics}

\par
\noindent{\textbf{Image quality.}}
We evaluate the quality of our generated images using CLIP Score (CS), Inception Score (IS), Blind/Referenceless Image Spatial Quality Evaluator (BRISQUE) \cite{mittal2012no} and Natural Image Quality Evaluator (NQIE) \cite{mittal2012making}. 
The Inception Score is based on the diversity and predictability of the generated images.
CLIP Score uses a pre-trained CLIP model~\cite{clip} to measure the similarity between text and images.
Note that existing visual quality metrics such as FID cannot be used since the scenes generated by text-to-3D approaches do not exhibit the same underlying data distribution.
\par
\noindent{\textbf{Multiview Consistency.}}
Given a sequence of rendered images, we evaluate the multi-view consistency of our generated scene using Camera Error (CE), Depth Error (DE), and flow-warping error (FE) metrics.
Motivated by~\cite{scenescape,chen2023scenedreamer}, we use COLMAP~\cite{schoenberger2016sfm}, a reliable SfM technique, to compute the camera trajectory and the sparse 3D point cloud. CE is computed by comparing the difference between the predicted trajectory and the given trajectory, and DE is computed by comparing the difference between the sparse depth map obtained by COLMAP and the estimated depth map.
In addition, to account for temporal consistency, we follow~\cite{lai2018learning} and use RAFT~\cite{teed2020raft} to compute FE.


\input{table/compare_text_to_scene}

\input{table/compare_to_video}
\input{table/compare_to_pano}
\input{figure/video_compare}
\input{table/ablation_core}
\input{figure/3d_res}

\subsection{Comparisons}
\noindent{\textbf{Baselines.}}
Since there are only a few baselines directly related to our approach, we also take into account some methods with similar capabilities and construct their variants for comparison. 
Specifically, the following three categories of methods are included:
\begin{itemize}
    \item \textbf{Text-to-Scene.} There exist techniques~\cite{scenescape, text2room} that generate 3D meshes iteratively by employing warping and inpainting processes, allowing for direct comparisons with our proposed methods. Moreover, image-guided 3D generation methods~\cite{Infinitenature-zero, 3D_Cinemagraphy, pixelsynth} are also available, wherein initial images can be produced using a T2I model. Subsequently, their pipeline can be used to generate 3D scenes, enabling a comparison against our approach. We comprehensively evaluate these methods based on the previously introduced 3D consistency and visual quality metrics.
    \item \textbf{Text-to-Video.} Some recent text-driven video generation methods~\cite{VideoFusion, Gen-2} can also generate similar 3D scene walkthrough videos. Since it is not supported to explicitly control the camera motion in the video generation methods, we only evaluated them in terms of visual quality and temporal consistency.
    \item \textbf{Text-to-Panorama.} This task generates perspective images covering the panoramic field of view, which is challenging to ensure consistency in the overlapping regions. We have selected two related methods~\cite{chen2022text2light, tang2023mvdiffusion} for comparisons.

\end{itemize}

\par
\noindent{\textbf{Comparison to Text-to-Scene Methods.}}
To generate the scenes, we use a set of test-specific prompts covering descriptions of indoor, outdoor and unreal scenes.
Each prompt generates an image sequence of 100 frames, and for a fair comparison, we set a fixed random seed. 
After that, we compute the metrics proposed in~\cref{sec:evaluation_metrics} on the generated image sequences and evaluate the effectiveness of the method.
As shown in~\cref{tab:campare_to_scene}, our method outperforms the mesh-based iterative generation methods in several metrics, especially for outdoor scenes. 
The quality of their generation results relies heavily on the generative and geometric prior and degrades over time due to error accumulation.
In addition, their use of a mesh to represent the scene makes it difficult to represent intense depth discontinuities, which are common in outdoor scenes.
Our method, on the other hand, adopts hybrid NeRF as the scene representation, which can cope with complex scenes, and our rectification mechanism can mitigate the effect of accumulated errors caused by inaccurate prior signals.


\par
\noindent{\textbf{Comparison to Text-to-Video Methods.}}
For comparison with the text-to-video model, we used the same collection of prompts as input to the model and generated 1,200 video clips.
We used the same metrics to evaluate the 3D consistency and visual quality of the videos generated by the T2V model and our rendered videos.
As shown in~\cref{tab:compare_to_video}, our method significantly outperforms the T2V model on all metrics, proving the effectiveness of our method.
The T2V model learns geometry and appearance prior by training on a large video dataset, but it lacks a unified 3D representation, making it difficult to ensure multi-view consistency of the generated content, as can be observed~\cref{fig:video_result}.

\par
\noindent{\textbf{Comparison to Text-to-Panorama Methods.}}
We evaluate the methods~\cite{tang2023mvdiffusion, chen2022text2light} on visual quality.~\cref{tab:compare_to_pano} and~\cref{fig:360} present the quantitative and qualitative evaluations, respectively.
From the results, it can be seen that the results of previous methods can be inconsistent at the left and right boundaries, while our method, although not specifically designed for panorama generation, produces multiple views with cross-view consistency.

\par
\noindent{\textbf{3D Results.}}
In~\cref{fig:3d_res}, we show the 3D results reconstructed by our method. The 3D mesh is extracted by the marching cube algorithm~\cite{marching}.
Additionally, we can reconstruct high-quality point clouds using colmap~\cite{schoenberger2016sfm} by inputting the rendered image collection, which further demonstrates the superior 3D consistency of the generated view results.

\subsection{Ablation Study}
To further analyze the proposed methodology, we performed several ablation studies to evaluate the effectiveness of each module.
More ablation studies can be found in our supplementary material.

\noindent{\textbf{Effectiveness of Unified Representations.}}
To validate our necessity to construct a unified 3D representation, we remove it from our pipeline.
At this time, our approach degenerates to the previous paradigm of warping-inpainting.
As shown in~\cref{tab:ablation_core}, the quality of the generated scenes degrades in DE and CE metrics due to the lack of global 3D consistency constraints.

\par
\noindent{\textbf{Effectiveness of Generative Refinement.}}
To validate the effectiveness of our proposed generative refinement, we ablate the modules in our approach, whereby the novel view obtained through volume rendering will be updated directly into the supporting database for subsequent incremental training.
The results in~\cref{tab:ablation_core} show that this can lead to a significant degradation in the quality of the generated scene.
We argue that the reason for this is that the quality of novel views generated by NeRF training on sparse views tends to be inferior, with notable blurring and artifacts.
Therefore, adding this data for optimizing 3D scenes would lead to continuous degradation of the quality of the generated scenes.

\par
\noindent{\textbf{Effectiveness of Consistency Regularization.}}
To verify the validity of our regularization loss, we ablate this loss and generate scenes to compute the relevant metrics.
As shown in ~\cref{tab:ablation_core}, adding this loss further improves the 3D consistency of the generated scenes.
Though we explicitly inject 3D information into the refining process, its output still shows some inconsistent results in several scenes.
Therefore, to further improve the quality of the generated new views, we perform test-time optimization through this regularization term to constrain the consistency between local and global representations.

%% file: figure/figure_result.tex
\begin{figure*}[t!]
    \centering
    \includegraphics[width=1.0\textwidth]{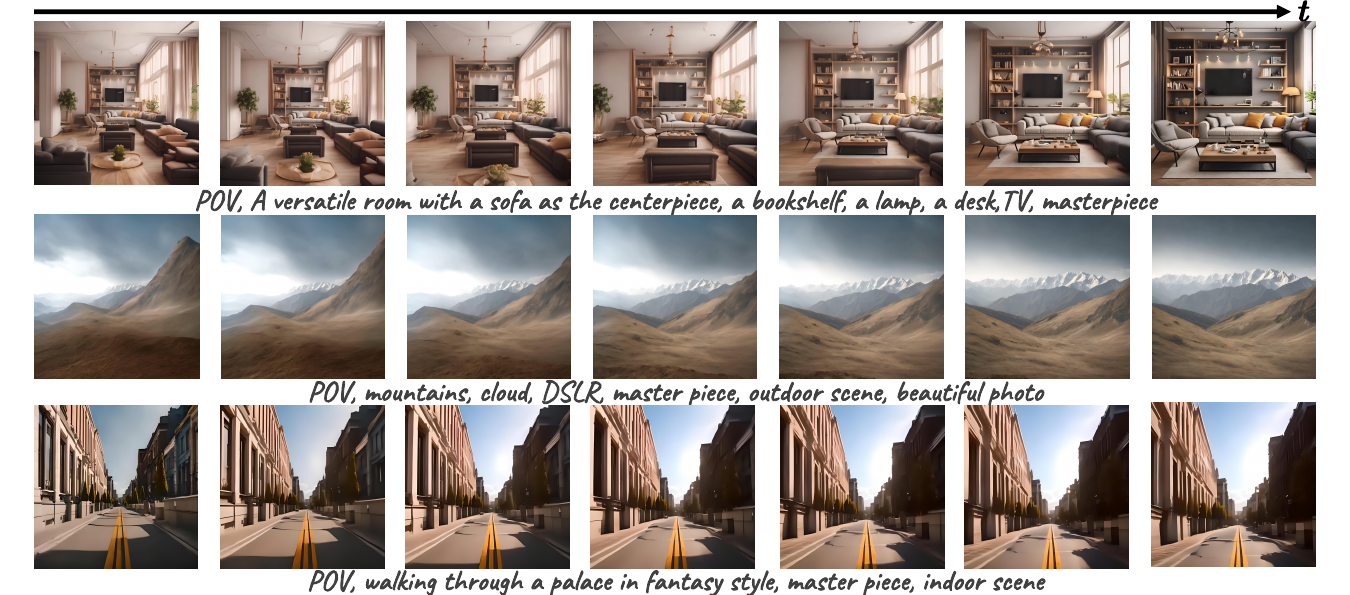}
    \vspace{-6mm}
    \caption{\textbf{Quantitative Results.} From our results, it can be seen that our approach produces high-fidelity scenes with stable 3D consistency in indoor scenes, outdoor scenes, and unreal-style scenes. More high-resolution results can be found in the supplementary material.}
    \vspace{-4mm}
    \label{fig:result}
\end{figure*}

%% file: figure/text2360.tex
\begin{figure}[t!]
    \centering
    \includegraphics[width=0.5\textwidth]{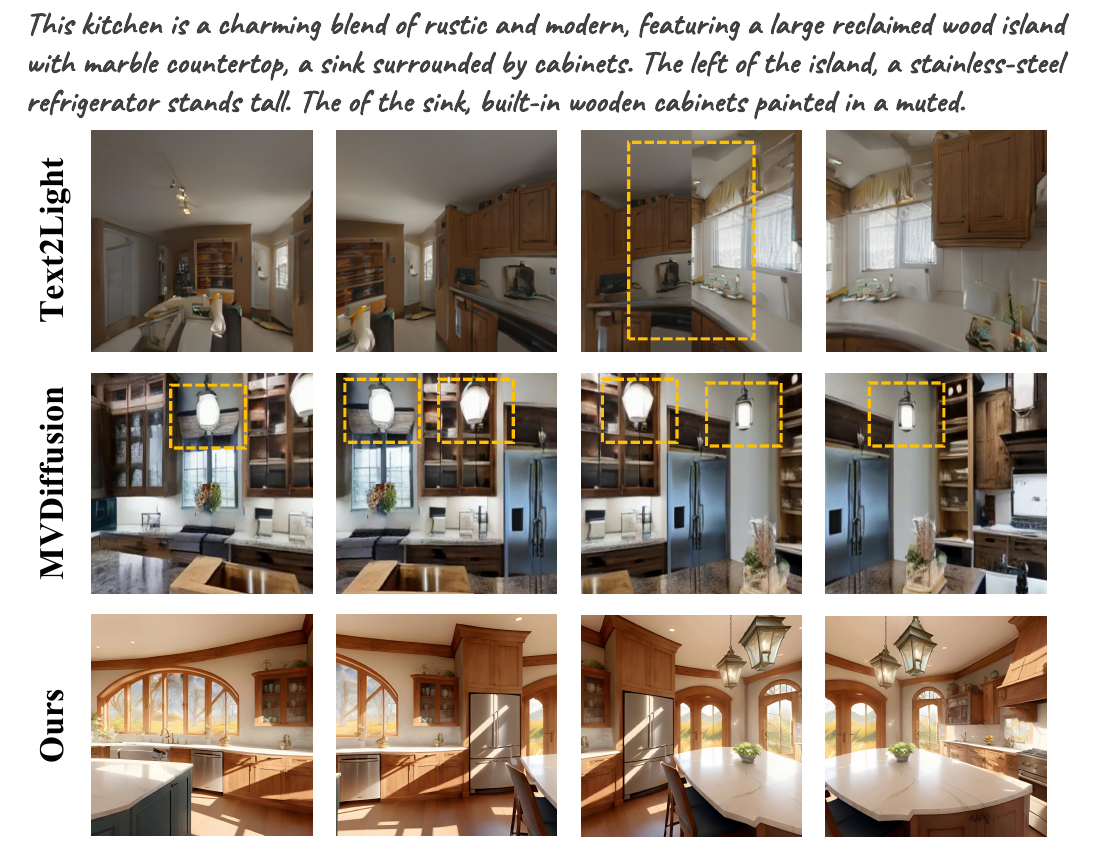}
    \vspace{-4mm}
    \caption{\textbf{Comparison with text-to-panorama methods}. It can be seen that although our method is not trained on panoramic data, it can also generate multiple views with cross-view consistency.}
    \vspace{-6mm}
    \label{fig:360}
\end{figure}

%% file: table/compare_text_to_scene.tex
\begin{table}[t!]
\centering
\resizebox{\columnwidth}{!}{%

\begin{tabular}{l|c|ccc|cccc}
\toprule
                     &             & \multicolumn{3}{c|}{\textbf{3D Consistentcy}}  & \multicolumn{4}{c}{\textbf{Visual Quality}}                     \\ \cmidrule(l){3-9} 
\multirow{-2}{*}{\textbf{Method}} &
  \multirow{-2}{*}{\textbf{\begin{tabular}[c]{@{}c@{}}3D\\ Representation\end{tabular}}} &
  \cellcolor[HTML]{FEF1F1}\textbf{DE$\downarrow$} &
  \cellcolor[HTML]{FEF1F1}\textbf{CE$\downarrow$} &
  \cellcolor[HTML]{F0FBEF}\textbf{SfM rate$\uparrow$} &
  \cellcolor[HTML]{F0FBEF}\textbf{CS$\uparrow$} &
  \cellcolor[HTML]{FEF1F1}\textbf{BRISQUE$\downarrow$} &
  \cellcolor[HTML]{FEF1F1}\textbf{NIQE$\downarrow$} &
  \cellcolor[HTML]{F0FBEF}\textbf{IS$\uparrow$} \\ \midrule
Inf-Zero~\cite{Infinitenature-zero} & -           & -             & 1.189          & 0.38          & -              & \textbf{21.43} & 5.85          & 2.34          \\
3DP~\cite{3dp}                  & LDI\&Mesh   & 0.42          & 0.965          & 0.47          & -              & 29.95          & 5.84          & 1.75          \\
PixelSynth~\cite{pixelsynth}           & Point Cloud & 0.36          & 0.732          & 0.52          & -              & 36.74          & 4.98          & 1.28          \\ \midrule
ProlificDreamer~\cite{Prolificdreamer}      & NeRF        & -             & -              & -             & 23.41          & 27.97          & 6.75          & 1.21          \\
Text2Room~\cite{text2room}            & Mesh        & 0.24          & 0.426          & 0.63          & 28.15          & 28.37          & 5.46          & 2.19          \\
Scenescape~\cite{scenescape}           & Mesh        & 0.18          & 0.394          & 0.76          & 28.84          & 24.54          & 4.78 & 2.23          \\
\rowcolor[HTML]{EFEFEF} 
Ours                 & NeRF        & \textbf{0.13} & \textbf{0.176} & \textbf{0.89} & \textbf{29.97} & 23.64          & \textbf{4.66}          & \textbf{2.62} \\ \bottomrule
\end{tabular}
}
\vspace{-2mm}
\caption{\textbf{Comparison with text-to-scene methods.} We compare our approach with two categories of approaches, \ie, pure text-driven 3D generation and text-to-image generation followed by 3D scene generation. Metrics on 3D consistency and visual quality are illustrated.}
\label{tab:campare_to_scene}
\vspace{-2mm}
\end{table}

%% file: table/compare_to_video.tex
\begin{table}[t!]
\centering
\resizebox{\columnwidth}{!}{%
\begin{tabular}{l|ccccc}
\toprule
\textbf{Method} &
  \cellcolor[HTML]{FEF1F1}\textbf{FE$\downarrow$} &
  \cellcolor[HTML]{F0FBEF}\textbf{CS$\uparrow$} &
  \cellcolor[HTML]{FEF1F1}\textbf{BRISQUE$\downarrow$} &
  \cellcolor[HTML]{FEF1F1}\textbf{NIQE$\downarrow$} &
  \cellcolor[HTML]{F0FBEF}\textbf{IS$\uparrow$} \\ \midrule
VideoFusion~\cite{VideoFusion} &
  0.039 &
  23.54 &
  27.39 &
  5.94 &
  2.21 \\
GEN-2~\cite{Gen-2} &
  0.032 &
  27.54 &
  25.65 &
  5.24 &
  2.38 \\
\rowcolor[HTML]{EFEFEF} 
Ours &
  \textbf{0.028} &
  \textbf{29.95} &
  \textbf{23.53} &
  \textbf{4.70} &
  \textbf{2.69} \\ \bottomrule
\end{tabular}%
}
\vspace{-2mm}
\caption{\textbf{Comparision with text-to-video methods}. Metrics on flow warping error (FE) and visual quality are illustrated.}
\label{tab:compare_to_video}
\vspace{-2mm}
\end{table}

%% file: table/compare_to_pano.tex
\begin{table}[t!]
\centering
\resizebox{\columnwidth}{!}{%
\begin{tabular}{l|cccc}
\toprule
\textbf{Method} &
  \cellcolor[HTML]{F0FBEF}\textbf{CS$\uparrow$} &
  \cellcolor[HTML]{FEF1F1}\textbf{BRISQUE$\downarrow$} &
  \cellcolor[HTML]{FEF1F1}\textbf{NIQE$\downarrow$} &
  \cellcolor[HTML]{F0FBEF}\textbf{IS$\uparrow$} \\ \midrule
Text2Light~\cite{CLIP_Mesh}       & 26.16          & 49.26          & 6.15          & 2.54          \\
MVDiffusion~\cite{dreamfusion}     & 27.25          & 31.54          & 5.47          & 2.76          \\
\rowcolor[HTML]{EFEFEF} 
Ours            & \textbf{28.12} & \textbf{24.15} & \textbf{4.96} & \textbf{2.79} \\ \bottomrule
\end{tabular}
}
\vspace{-2mm}
\caption{\textbf{Comparision with text-to-panorama methods.} We compare our method with recent text-driven 3D generation methods~\cite{tang2023mvdiffusion, chen2022text2light}. Metrics on visual quality are illustrated.}
\label{tab:compare_to_pano}
\vspace{-2mm}
\end{table}

%% file: figure/video_compare.tex
\begin{figure}[t!]
    \centering
    \includegraphics[width=0.5\textwidth]{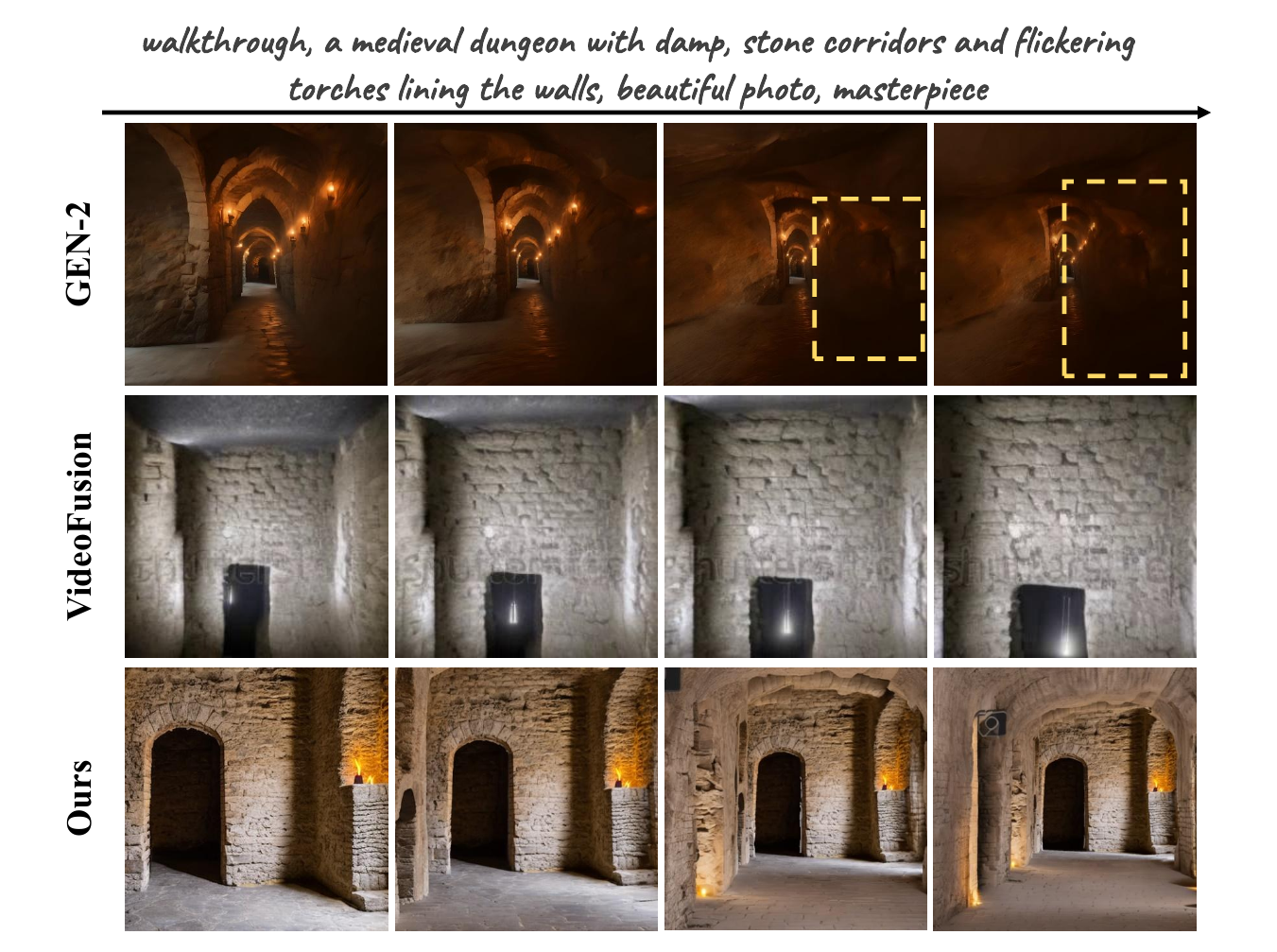}
    \vspace{-4mm}
    \caption{\textbf{Comparison with text-to-video methods.}
    Blur artifacts and temporally inconsistent frames occur in the text-to-video methods because of the lack of global 3D representation.} 
    \vspace{-6mm}
    \label{fig:video_result}
\end{figure}

%% file: table/ablation_core.tex
\begin{table}[t!]
\centering
\resizebox{\columnwidth}{!}{%
\begin{tabular}{l|cccccc}
\toprule
\textbf{Method} &
  \cellcolor[HTML]{FEF1F1}\textbf{DE$\downarrow$} &
  \cellcolor[HTML]{FEF1F1}\textbf{CE$\downarrow$} &
  \cellcolor[HTML]{F0FBEF}\textbf{SfM rate$\uparrow$} &
  \cellcolor[HTML]{F0FBEF}\textbf{CS$\uparrow$} &
  \cellcolor[HTML]{FEF1F1}\textbf{BRISQUE$\downarrow$} &
  \cellcolor[HTML]{FEF1F1}\textbf{NIQE$\downarrow$} \\ \midrule
\rowcolor[HTML]{EFEFEF} 
Full Model  & \textbf{0.13} & \textbf{0.176} & \textbf{0.89} & \textbf{29.97} & \textbf{26.18} & \textbf{6.54} \\
\rowcolor[HTML]{FFFFFF} 
W/o UR & 0.46          & 0.764          & 0.41          & 22.71          & 27.95          & 5.81          \\
W/o RM        & 0.59          & 0.981          & 0.46          & 22.12          & 29.64          & 5.75          \\
W/o CR & 0.19          & 0.254          & 0.78          & 28.14          & 27.16          & 6.12          \\ \bottomrule
\end{tabular}%
}
\vspace{-2mm}
\caption{\textbf{Ablations.} For brevity, we use UR, GRM, CR to denote \textit{Unified Representation}, \textit{Generative Refinement Model} and \textit{Consistency Regularization}, respectively.}
\label{tab:ablation_core}
\vspace{-6mm}
\end{table}

%% file: figure/3d_res.tex
\begin{figure}[t]
    \centering
    \includegraphics[width=0.48\textwidth]{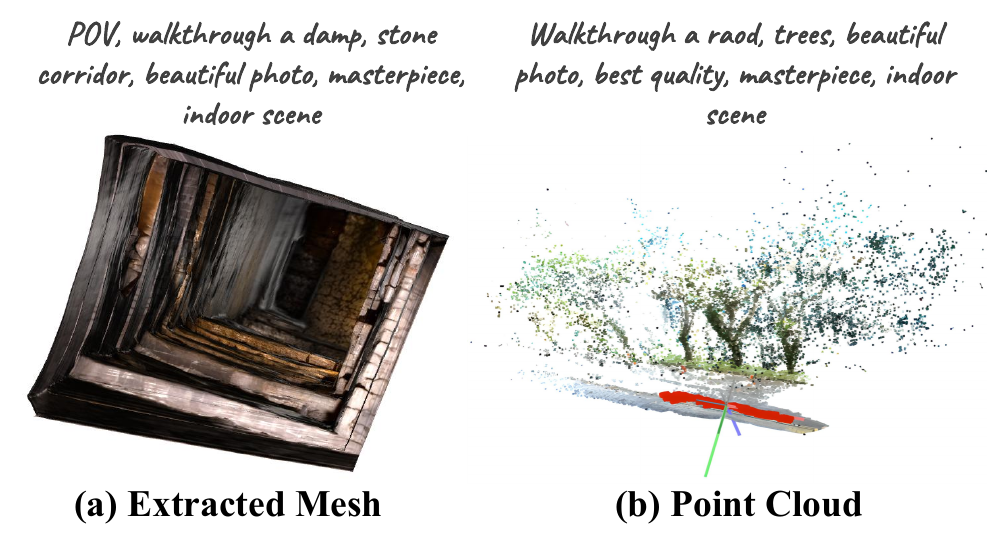}
    \vspace{-3mm}
    \caption{\textbf{Reconstructed 3D Results.} (a) The 3D mesh extracted by marching cube algorithm, and (b) the point cloud obtained after the reconstruction using COLMAP~\cite{marching}. Our reconstruction results show that our methods can generate scenes with satisfactory 3D consistency.}
    \vspace{-6mm}
    \label{fig:3d_res}
\end{figure}

%% file: sec/6_conclution.tex
\section{Conclusion}
\label{sec:conclusion}

This paper presents a new framework, which employs the tri-planar feature-based neural radiation field as a unified 3D representation and provides a unified solution for text-driven indoor and outdoor scene generation and the output supports navigation with arbitrary camera trajectories.
Our method fine-tunes a scene-adapted diffusion model to correct the generated new content to mitigate the effect of cumulative errors while synthesizing extrapolated content.
Experimental results show that our method can produce results with better visual quality and 3D consistency compared to previous methods.